\documentclass[11pt]{article}
\usepackage{verbatim}
\usepackage[preprint]{acl}

\usepackage{times}
\usepackage{latexsym}
\usepackage{multirow}
\usepackage{array}
\usepackage{makecell}
\usepackage{threeparttable}
\usepackage{enumitem}
\usepackage{microtype}
\usepackage[T1]{fontenc}

\usepackage[utf8]{inputenc}

\usepackage{microtype}

\usepackage{inconsolata}

\usepackage{graphicx}
\usepackage{caption}
\usepackage{float}
\usepackage{booktabs} 
\usepackage{hyperref}

\usepackage{stfloats}

\usepackage{pifont}
\usepackage{colortbl}
\usepackage{tabularx}
\usepackage{adjustbox}

\title{GlobeAudio: A Multilingual Multicultural Benchmark for Naturalistic Evaluation of Large Audio-Language Models}

\author{Ryner Tan, Wenxuan Zhang \\
        Singapore University of Technology and Design \\
        \texttt{ryner\_tan@mymail.sutd.edu.sg}, \texttt{wxzhang@sutd.edu.sg}}

\begin{document}

\maketitle

\begin{abstract}

Large Audio-Language Models (LALMs) integrate audio perception and language understanding within a unified framework, enabling a wide range of real-world applications. Despite recent advances, evaluation for LALMs remains heavily underspecified relative to real-world requirements: most lack true linguistic and cultural authenticity, while others fail to capture acoustic realism. To bridge this gap, we propose \textsc{GlobeAudio}, a multilingual and multicultural benchmark designed to evaluate naturalistic audio understanding. \textsc{GlobeAudio} consists of 5,637 multiple-choice questions across six typologically diverse languages, expertly crafted by native speakers grounded on naturally occurring audio. In order to do well, models must possess higher-level auditory reasoning skills and culturally grounded interpretation. We systematically evaluate representative closed-source and open-source LALMs, as well as cascaded ASR–LLM pipelines. Our experiments reveal substantial performance gaps under natural acoustic conditions, particularly for open-source models and low-resource languages. These findings highlight critical limitations of current LALMs and underscore the importance of naturalistic audio evaluation for future audio-language systems. \textsc{GlobeAudio} can be found at \url{https://huggingface.co/datasets/iNLP-Lab/GlobeAudio}.

\end{abstract}

\section{Introduction}
Large Audio-Language Models (LALMs) are an emerging class of systems designed to comprehend, process, and generate responses directly from audio inputs \cite{lakhotia-etal-2021-generative,chu2024qwen2audiotechnicalreport,fang2025llamaomni,tang2024salmonn,liu2025voxtral,openai2024gpt4ocard}. Unlike traditional speech processing models, which typically specialize in individual tasks such as automatic speech recognition (ASR) \cite{radford2022whisper,sekoyan2025canary1bv2parakeettdt06bv3efficient,king2025flavorsmoonshinetinyspecialized,shakhadri2025sambaasrstateoftheartspeechrecognition}, LALMs unify audio processing and language understanding, enabling tasks like spoken QA and speech translation for diverse applications such as virtual assistants and educational tools \cite{su2025audiolanguagemodelsaudiocentrictasks,NEURIPS2023_3a2e5889}. Driven by their rapid adoption, the community has introduced evaluation suites ranging from those that consolidate and standardise traditional audio tasks \cite{wang2025audiobenchuniversalbenchmarkaudio,heigold2025massive} to broader benchmarks assessing multimodal reasoning across heterogeneous inputs \cite{sakshi2024mmaumassivemultitaskaudio,yang2024airbenchbenchmarkinglargeaudiolanguage}.

\begin{figure}[t!]
    \includegraphics[width=\columnwidth]{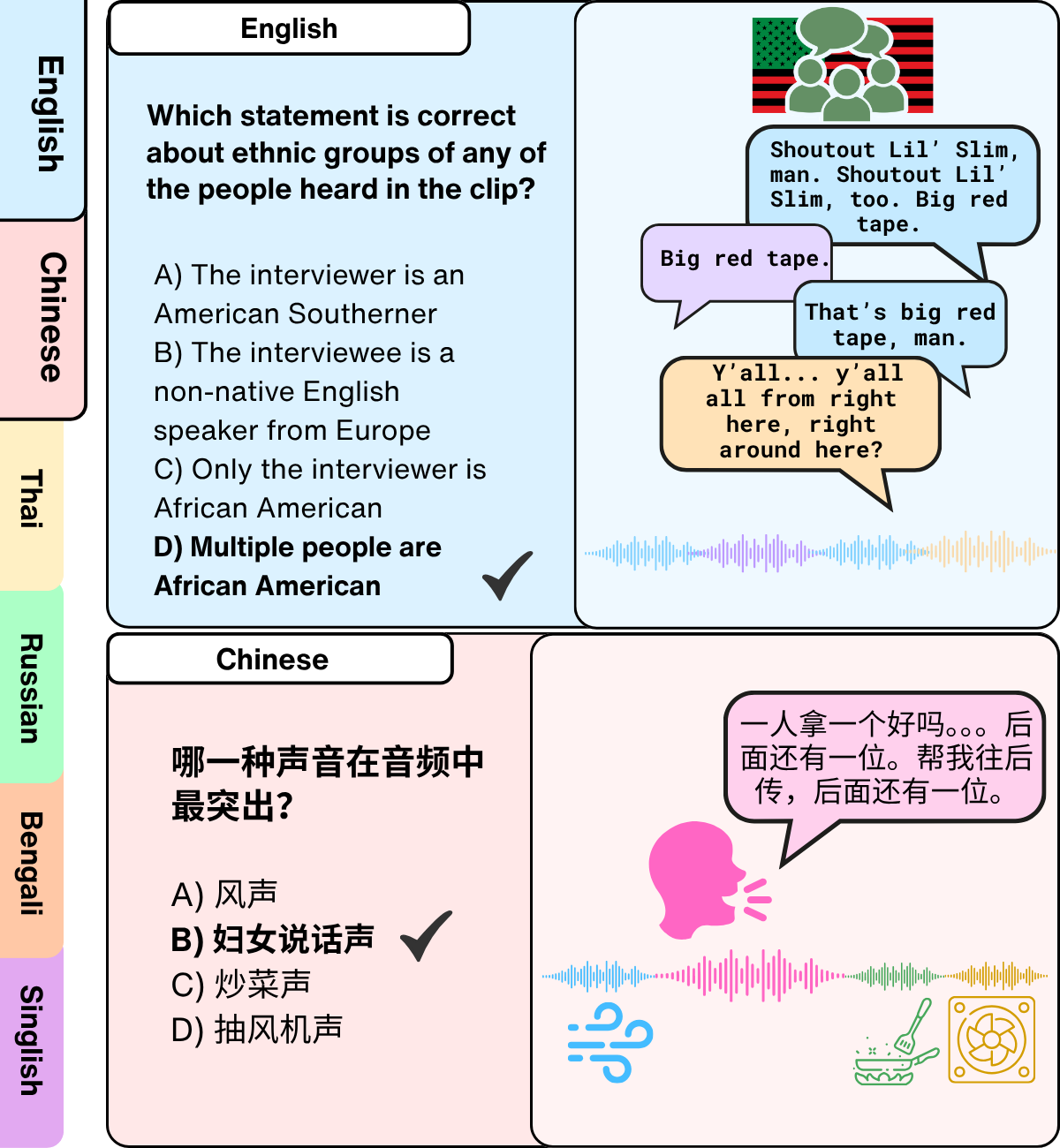}
    \caption{Examples of \textsc{GlobeAudio}, which require models to first unpack compound auditory tasks through multiple reasoning steps.
    }
    \label{fig:qn-example}
    
\end{figure}

\begin{table*}[t!]
  \centering
  \resizebox{0.9\linewidth}{!}{
  \begin{tabular}{lcccc}
    \toprule
    \textbf{Dataset}
      & \textbf{Multilingual}
      & \textbf{Multicultural}
      & \textbf{Natural / Assesses Prosody}
      & \textbf{Compound Tasks} \\
    \midrule
    AudioBench \cite{wang2025audiobenchuniversalbenchmarkaudio}    & \ding{55} & \ding{55} & \ding{51} & \ding{55} \\
    MSEB  \cite{heigold2025massive}         & \ding{51} & \ding{51} & \ding{55} & \ding{55} \\
    MMAU    \cite{sakshi2024mmaumassivemultitaskaudio}       & \ding{55} & \ding{55} & \ding{51} & \ding{51} \\
    AIR-Bench \cite{yang2024airbenchbenchmarkinglargeaudiolanguage}     & \ding{55} & \ding{55} & \ding{55} & \ding{51} \\
    Fleurs-SLU \cite{schmidt2025fleursslumassivelymultilingualbenchmark}    & \ding{51} & \ding{55} & \ding{55} & \ding{55} \\
    mSTEB  \cite{beyene2025mstebmassivelymultilingualevaluation}        & \ding{51} & \ding{55} & \ding{55} & \ding{55} \\
    AudioMarathon \cite{he2025audiomarathoncomprehensivebenchmarklongcontext} & \ding{55} & \ding{55} & \ding{55} & \ding{51} \\
    SSEU-Bench \cite{yin2025largeaudiolanguagemodels}    & \ding{55} & \ding{55} & \ding{55} & \ding{51} \\
    \midrule
    \textbf{GlobeAudio} & \textbf{\ding{52}} & \textbf{\ding{52}} & \textbf{\ding{52}} & \textbf{\ding{52}} \\
    \bottomrule
  \end{tabular}}
  \caption{Dataset comparison of GlobeAudio with other audio benchmarks in modelling real-world scenarios.}
  \label{tab:datasetcomparison}
\end{table*}

Despite these advances, evaluation for LALMs remains vastly underspecified in several important respects relative to real-world deployment requirements. Concretely, they remain predominantly English-centric and lack true linguistic and cultural authenticity. Though some multilingual benchmarks \cite{schmidt2025fleursslumassivelymultilingualbenchmark, beyene2025mstebmassivelymultilingualevaluation} are scalable and largely broaden language coverage, their reliance on automated translation compromises multicultural validity by flattening culturally grounded references.

In addition, current evaluations fail to capture acoustic realism by focusing primarily on lexical content \cite{sakshi2024mmaumassivemultitaskaudio, schmidt2025fleursslumassivelymultilingualbenchmark,beyene2025mstebmassivelymultilingualevaluation, heigold2025massive} thus largely overlooking prosodic cues critical to real-world communication, or relying on semi-scripted speech \cite{he2025audiomarathoncomprehensivebenchmarklongcontext,yin2025largeaudiolanguagemodels,lee2024speechmassivemultilingualspeechdataset,conneau2022xtremesevaluatingcrosslingualspeech} which fails to capture the acoustic and sociolinguistic variability of real-world contexts. Consequently, the scope of LALM evaluation remains reduced and constrained to decomposed, atomic tasks \cite{schmidt2025fleursslumassivelymultilingualbenchmark, heigold2025massive}, ignoring higher-level pragmatics and cultural context inherent to the real-world.

To bridge these critical gaps summarised in Table \ref{tab:datasetcomparison}, we introduce \textsc{GlobeAudio}, a multilingual and multicultural benchmark for naturalistic audio understanding. We capture the wide sociolinguistic and acoustic variability of real-world communication by curating unscripted, naturally occurring audio clips directly from online media across diverse contexts. Consequently, rather than relying on automated translations, native speakers deeply rooted in local culture authored all multiple-choice questions with carefully designed distractors that require fine-grained audio and cultural understanding beyond transcription. Finally, we ensure the highest data fidelity by employing a rigorous two-stage cross-review and adjudication process to resolve ambiguities, yielding an exceptionally high inter-annotator consensus at \textbf{95.5\%}. In total, \textsc{GlobeAudio} covers six languages spanning high to low-resource settings and diverse language families, offering a total of 5,637 expertly crafted multiple-choice questions in total as shown in Figure \ref{fig:qn-example}.

Consequently, we conduct a comprehensive evaluation of representative LALMs as well as cascaded pipelines (ASR $\rightarrow$ text LLM) under controlled settings. Our results reveal significant gaps in their compound auditory reasoning capabilities, demonstrating that naturalistic audio conditions remain a major blind spot for state-of-the-art models. We discover that this gap is especially severe in open-source models which trail considerably behind their closed-source counterparts, and in low-resource languages such as Thai and Bengali. Through targeted ablations and extensive analysis, we further demonstrate the high fidelity of \textsc{GlobeAudio} and reveal a surprising alignment behaviour: LALMs perform better when evaluated with questions and transcripts in the source language, instead of in English which is a comparatively higher resource language.

In summary, our contributions are threefold:
\begin{itemize}[itemsep=1pt]
    
    \item  We introduce \textsc{GlobeAudio}, a multilingual, multicultural benchmark for naturalistic audio understanding, comprising 5,637 human-authored and rigorously verified MCQs across 6 languages and diverse real-world contexts.
    \item We extensively benchmark various state-of-the-art LALMs, spanning closed-source, open-source models across various model families, and reveal substantial performance gaps, particularly for open-source models and low-resource languages.
    \item Through targeted ablations and cross-lingual studies, we uncover a counter-intuitive but consistent advantage of source-language evaluation over English translation, and the linguistic–cultural grounding in shaping LALM performance.
    
\end{itemize}

\begin{figure*}[t]
    \centering
    \includegraphics[width=0.98\textwidth]{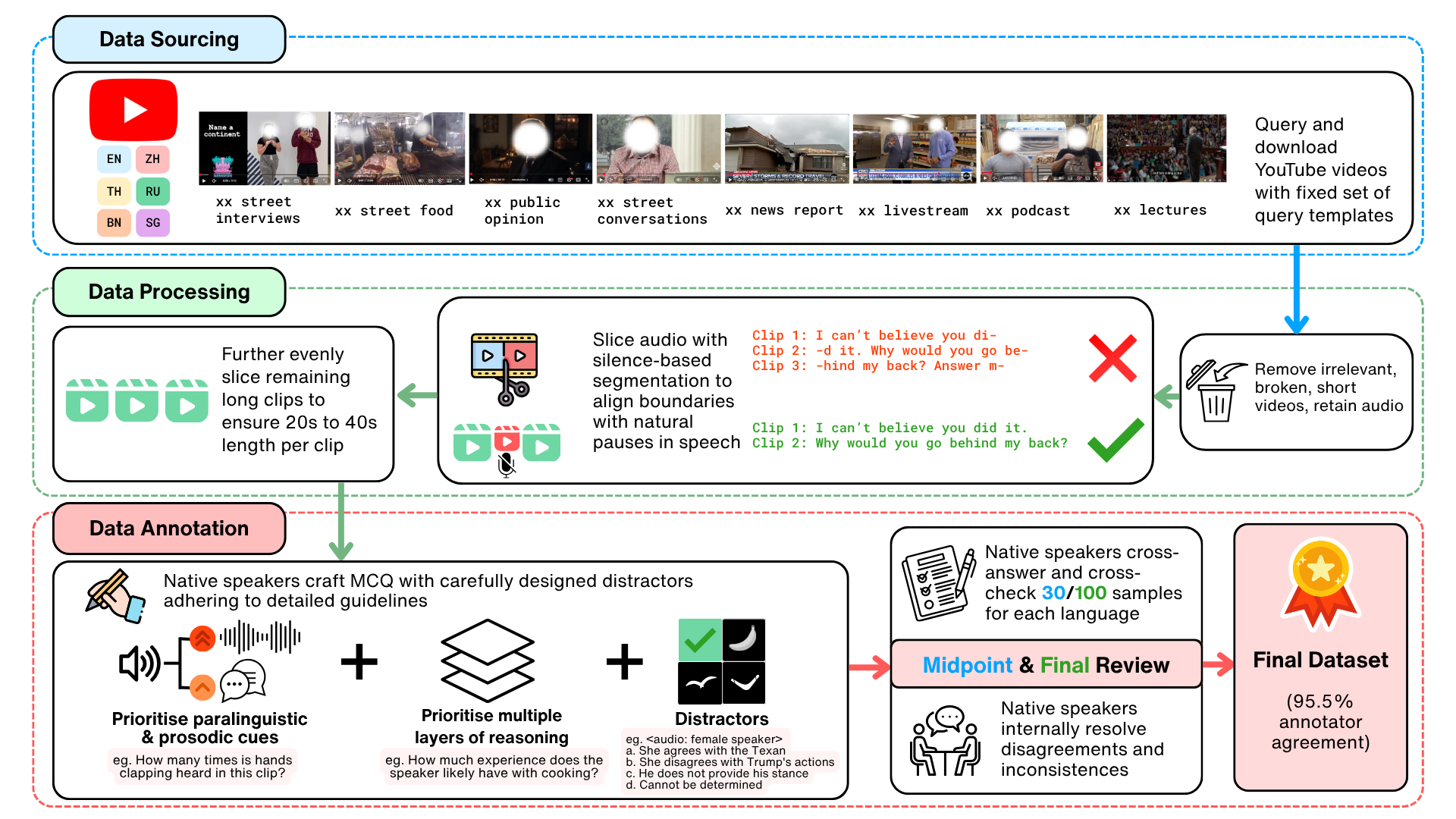}
    \caption{\textsc{GlobeAudio}'s data preparation pipeline, from curation to rigorous annotation and iterative cross-annotator quality checks, culminating in a high-fidelity dataset.}
    \label{fig:datapipeline}
\end{figure*}

\section{\textsc{GlobeAudio} Dataset}

\subsection{Design Considerations}
To provide a realistic reflection of real-world performance when LALMs are exposed to the acoustic and sociolinguistic conditions encountered \textit{in-the-Wild}, we design \textsc{GlobeAudio} with three core principles:

\begin{itemize}
    \item  \textit{Multilingual and Multicultural Coverage}: Real-world speech understanding is inherently shaped by linguistic diversity and culturally grounded usage \cite{alam2025everydaymmqamultilingualmultimodalframework}. To ensure that evaluation extends beyond English-centric or culturally homogenized settings, \textsc{GlobeAudio} is constructed by native speakers with lived experience in the target languages and cultures.

    \item \textit{In-the-Wild Audio Sources}: Spoken language in real-world environments is often characterized by background noise, prosody, overlapping speakers, informal delivery, and unconstrained recording conditions \cite{NAGRANI2020101027}. To capture this variability, \textsc{GlobeAudio} is built from naturally occurring audio clips drawn from online media spanning both formal and informal contexts. This design choice intentionally departs from clean or semi-scripted speech, exposing models to the realistic acoustic complexity that deployed LALMs must routinely handle.
    
    \item \textit{Compound Task Design}: 
    Naturalistic audio understanding demands higher-level skills like discourse tracking, pragmatic inference and culturally grounded interpretation. To this end, \textsc{GlobeAudio} adopts compound question formulations where answering each question typically requires several layers of reasoning over the audio content, discouraging reliance on superficial cues or isolated subtasks, but instead probing holistic auditory comprehension.
    
\end{itemize}

\begin{table*}[t]
\centering
\small
\begin{threeparttable}
\begin{tabular}{@{} l l l l l r r r @{}} 
\toprule
& & \multicolumn{3}{c}{\textbf{Linguistic Properties}} & \multicolumn{3}{c}{\textbf{Dataset Statistics}} \\
\cmidrule(lr){3-5} \cmidrule(lr){6-8}

\textbf{Language (Code)} & \textbf{CC Size} & \textbf{Isochrony} & \textbf{Genus} & \textbf{Script} & \textbf{Clip length} & \textbf{QA length (Q/A)} & \textbf{Total} \\

\midrule

English (en)  & 42.60 & High, Str   & Germanic     & Latin    & 25.72 & 53.8 / 22.9 & 1,274 \\
Russian (ru)  & 6.15  & High, Str   & Slavic       & Cyrillic & 25.69 & 37.8 / 22.2 & 924 \\
Chinese (zh)  & 4.99  & Mid, Syl    & Sinitic      & Hanzi    & 24.88 & 13.5 / 5.0  & 1,072 \\
Thai (th)     & 0.37  & Mid, Syl      & Tai          & Thai     & 24.82 & 29.3 / 13.0 & 1,145 \\
Bengali (bn)  & 0.10  & High, Syl   & Indo-Aryan   & Bengali  & 25.47 & 37.6 / 18.3 & 719 \\
Singlish (sg) & --    & Low, Syl    & -- & Latin    & 25.54 & 73.3 / 34.9 & 503 \\

\midrule

\textbf{Total} & & & & & \textbf{25.35} & \textbf{40.9 / 19.4} & \textbf{5,637} \\

\bottomrule
\end{tabular}
\begin{tablenotes}
\small
\vspace{2pt}
\item \textit{Isochrony:} Categorised by syllable complexity and rhythmic unit (\textit{Str}: stress-timed, \textit{Syl}: syllable-timed)
\end{tablenotes}
\caption{
    Dataset Composition of \textsc{GlobeAudio}, sorted by decreasing language resource level. We report weighted averages for clip length (in seconds) and QA length (in characters), with per-language count.}
\label{tab:datasetcomposition}

\end{threeparttable}
\end{table*}

\subsection{Language Selection and Considerations}
Guided by the above design principles, we aim to construct an evaluation set that meaningfully probes both linguistic and cultural variability in real-world speech understanding. Language selection is therefore driven by a combination of complementary criteria, including geographical coverage, resource availability, diversity in writing systems, and diversity in phonological characteristics.

Based on these considerations, we select six languages: \textbf{English} (United States), \textbf{Chinese} (China), \textbf{Thai} (Thailand), \textbf{Russian} (Russia), \textbf{Bengali} (India) and \textbf{Singlish} (Singapore). Collectively, this set spans multiple language families (Indo-European, Sino-Tibetan, Tai–Kadai, and creole varieties), covers a wide range of geographic regions, and reflects substantial variation in resource availability. As shown in Table~\ref{tab:datasetcomposition}, where we use CommonCrawl's Size (CC Size) as a proxy for the resource level, it covers from high-resource languages such as English, to comparatively much lower-resource settings such as Thai and Bengali. The selected languages also differ markedly in script systems and spoken characteristics: the inclusion of Singlish enables the evaluation of naturally occurring code-mixing and multilingual speech within a single variety, capturing linguistic phenomena that are common in real-world audio but rarely represented in existing benchmarks.

\subsection{Data Collection and Processing}
We present a full, detailed visualisation of our data preparation pipeline in Figure \ref{fig:datapipeline}, which we describe in detail below.

YouTube hosts a large and continuously updated collection of publicly accessible video content spanning a wide range of communicative contexts. We therefore source our raw data from YouTube using \texttt{yt-dlp}, an open-source audio and video downloader tool \footnote{GitHub: \url{https://github.com/yt-dlp/yt-dlp}} on a curated set of language-specific queries designed to capture both formal and informal speech. For each language, we retrieve 15 videos with each of the following query templates: \textit{\{"xx street food", "xx street interviews", "xx street conversations", "xx public opinion", "xx news report", "xx lectures", "xx livestream", "xx podcast"\}}, where x refers to the corresponding country of the selected languages.

From the retrieved videos, we discard visual content and retain only the audio streams for further processing. To obtain manageable yet contextually rich audio samples while avoiding mid-utterance truncation, we segment each full-length audio file using silence-based segmentation, aligning clip boundaries with natural pauses in speech. The silence-based audio slicer is open-sourced on GitHub.\footnote{\url{https://github.com/openvpi/audio-slicer}}. This procedure yields short audio clips with durations ranging from 20 to 40 seconds, which balances sufficient discourse context with practical annotation and evaluation constraints.

\subsection{Data Annotation}

Data annotation is performed exclusively by native speakers in their first language, ensuring high linguistic fidelity and culturally appropriate interpretation. For each audio clip, these annotators construct a four-option multiple-choice question (MCQ) with one correct answer and three carefully designed distractors. We adopt the MCQ format over open-ended ones since it enables scalable, unambiguous automatic evaluation while preserving difficulty.
Concretely, annotators are instructed to prioritize questions that rely on information conveyed through the audio signal itself, such as and not limited to pragmatic meaning and paralinguistic cues, or culturally grounded questions. Furthermore, they are encouraged to craft ones that require multi-step reasoning over the audio, rather than direct fact retrieval. Annotators also design distractors to be semantically or acoustically confusable to discourage superficial elimination strategies. Finally, they do not annotate clips for which no meaningful question can be asked, upon our verification.

To ensure high annotation quality, we conduct a two-stage quality control process consisting of a midpoint review and a final review. In the midpoint review, we randomly sample 30 annotated items from each annotator and perform cross-checking among native speakers within the same language to identify annotation errors and inconsistencies at an early stage. Following completion, we conduct a final review by applying stratified random sampling to select an additional 500 items (100 per language), ensuring balanced representation across annotators. These samples are independently cross-checked and aligned in the same manner as the midpoint review. Across the reviewed samples, we observe an inter-annotator agreement of \textbf{95.5\%}. Disagreements are resolved through discussion and correction, resulting in a final dataset with verified and consistent gold labels as visualised in Figure \ref{fig:datapipeline}.

\subsection{Dataset Breakdown}
Finally, after data processing and quality control, \textsc{GlobeAudio} comprises a total of 5,637 multiple-choice questions spanning five languages and diverse language families. All audio clips are between 20 to 40 seconds long, with the mean clip length being 25.35s. A detailed breakdown is found in Table \ref{tab:datasetcomposition}.

Figure \ref{fig:qn-example} illustrates representative samples from our dataset, comprising of audio inputs and their corresponding multiple-choice questions, which require models to unpack compound auditory tasks through multiple reasoning layers. More examples can be found in Appendix Section \ref{sec:appendix-eg}.

\section{Experimental Setup}

\subsection{Model Selection}

For comprehensive evaluation on \textsc{GlobeAudio}, we select a diverse set of representative models spanning both open-source and closed-source LALMs, model families, architectures, and deployment trade-offs. In addition, we include cascaded Automatic Speech Recognition (ASR)–LLM pipelines as strong baselines for comparison against end-to-end LALMs.

\paragraph{Closed-source LALMs}
We consider two state-of-the-art LALMs, namely \texttt{Gemini 3.1 Pro Preview} \cite{gemini3.1promodelcard}, which is a frontier multimodal model with strong cross-modal reasoning capabilities, and \texttt{GPT-4o Audio} \cite{openai2024gpt4ocard}, which achieves state-of-the-art performance in multiple audio benchmarks including translation and transcription. We also include \texttt{Gemini 2.0 Flash} \cite{comanici2025gemini25pushingfrontier} as a latency and cost-optimized multimodal model, allowing for evaluation of trade-offs in real-world deployment scenarios.
\paragraph{Open-source LALMs}
We extensively evaluate seven open-source LALMs spanning a wide range of parameter scales, from smaller models like Alibaba's \texttt{Qwen2-Audio-7B-Instruct} \cite{chu2024qwen2audiotechnicalreport} to much larger models like Xiaomi's \texttt{MiMo-V2.5} \cite{xiaomimimov2.5}. We also include \texttt{Qwen3-Omni-Flash-2025-09-15} \cite{xu2025qwen3omnitechnicalreport}, the latest high-capacity Mixture-of-Experts model which is state-of-the-art on most existing audio and audio-visual benchmarks.

\paragraph{Cascaded ASR–LLM models}
In contrast to end-to-end LALMs, cascaded models decompose audio understanding into sequential components. We evaluate two such pipelines that first transcribe audio using a Whisper-based ASR system (\texttt{whisper-1}) \cite{radford2022whisper}, followed by downstream reasoning with text-only input. For the second component, we select \texttt{Qwen3-235B-A22B-Instruct-2507} \cite{yang2025qwen3technicalreport} which has 235B parameters and \texttt{Gemini 3.1 Pro Preview} \cite{gemini3.1promodelcard}, allowing for direct comparisons against their end-to-end counterparts under controlled conditions.
\subsection{Settings}
\paragraph{Inference}
Inference of all models is performed through API calling, with the exception of \texttt{Qwen2-Audio-7B-Instruct} which is performed on vLLM \footnote{\url{https://github.com/vllm-project/vllm}}, an open-source high-throughput inference engine for LLMs. In the event of API timeouts or invalid responses, we perform a maximum of 3 retries to ensure reproducibility.
\paragraph{Prompts}

Across all models, we employ a unified template detailed in Appendix Section \ref{sec: appendix-prompts}. For end-to-end open- and closed-source LALMs, the audio clip is fed directly into the model. For cascaded pipelines, the ASR-generated transcript is provided as context before prompting the LLM. To avoid introducing cross-lingual confounds, all prompts are in the source language.

\paragraph{Evaluation Metrics}
As \textsc{GlobeAudio} consists exclusively of MCQ items, model outputs are verifiable and evaluated on accuracy, comparing the predicted option label against the ground-truth.

\begin{table*}[t]
\centering
\begin{adjustbox}{max width=\textwidth}
\fontsize{9}{10.5}\selectfont
\setlength{\tabcolsep}{2.5pt}
\renewcommand{\arraystretch}{1}
\begin{tabular}{@{}l l c cccccc c@{}}
\toprule
\multirow{2}{*}{\textbf{Model}} & \multirow{2}{*}{\textbf{Modalities}} & \multirow{2}{*}{\textbf{Size}} & \multicolumn{6}{c}{\textbf{Language Accuracy (\%)}} & \multirow{2}{*}{\textbf{Avg.}} \\
\cmidrule(lr){4-9}
\noalign{\vspace{-2pt}}
& & & \textbf{en} & \textbf{ru} & \textbf{zh} & \textbf{th} & \textbf{sg} & \textbf{bn} & \\
\midrule \midrule
\multicolumn{10}{l}{\centerline{\textbf{Closed-source Models}}} \\
\midrule \midrule
GPT-4o Audio {\small \cite{openai2024gpt4ocard}} & $T, A \rightarrow T, A$ & -- & 61.62 & 70.24 & 50.75 & 45.07 & 76.74 & 56.33 & 60.13 \\
Gemini 2.0 Flash {\small \cite{gemini2flashmodelcard}} & $T, A \rightarrow T$ & -- & 64.13 & 77.81 & 55.13 & 60.35 & 74.75 & 68.15 & 66.72 \\
Gemini 3.1 Pro {\small \cite{gemini3.1promodelcard}} & $T, A \rightarrow T$ & -- & \textbf{78.26} & \textbf{90.04} & \textbf{70.34} & \textbf{81.57} & \textbf{90.06} & \textbf{85.81} & \textbf{82.68} \\
\midrule \midrule
\multicolumn{10}{l}{\centerline{\textbf{Open-source Models}}} \\
\midrule \midrule
Qwen2-Audio {\small \cite{chu2024qwen2audiotechnicalreport}} & $T, A \rightarrow T$ & 7B & 48.04 & 55.30 & 40.95 & 29.00 & 54.67 & 25.87 & 42.31 \\
MERaLiON 2 {\small \cite{he2024meralionaudiollmtechnicalreport}} & $T, A \rightarrow T$ & 10B & 53.38 & 60.82 & 48.79 & 45.15 & 69.38 & 56.33 & 55.64 \\
Audio Flamingo 3 {\small \cite{goel2025audioflamingo3advancing}} & $T, A \rightarrow T$ & 7B & 64.99 & 71.65 & 52.38 & 46.81 & 73.36 & 28.93 & 56.35 \\
Gemma 3n {\small \cite{gemma3nmodeloverview}} & $T, A \rightarrow T$ & 8B & 55.49 & 64.83 & 45.62 & 48.82 & 65.61 & 58.83 & 56.53 \\
Voxtral {\small \cite{liu2025voxtral}} & $T, A \rightarrow T$ & 24B & 59.65 & 68.51 & 51.77 & 46.99 & 75.35 & 45.34 & 57.94 \\
Qwen3-Omni {\small \cite{xu2025qwen3omnitechnicalreport}} & $T, A \rightarrow T, A$ & 30B & 71.35 & 82.25 & 63.34 & 65.15 & 80.32 & 70.24 & 72.11 \\
MiMo-V2.5 {\small \cite{xiaomimimov2.5}} & $T, A \rightarrow T$ & 310B & 65.54 & 68.51 & 57.60 & 43.23 & 75.94 & 55.63 & 61.08 \\
\midrule \midrule
\multicolumn{10}{l}{\centerline{\textbf{Cascade Systems}}} \\
\midrule \midrule
Whisper + Qwen 3 & $A \rightarrow T \rightarrow T$ & 1.5B + 30B & 65.23 & 78.03 & 51.40 & 60.52 & 74.55 & 58.41 & 64.69 \\
Whisper + Gemini 3.1 Pro & $A \rightarrow T \rightarrow T$ & 1.5B + -- & 69.78 & 80.84 & 56.81 & 71.79 & 82.50 & 70.51 & 72.04 \\
\bottomrule
\end{tabular}
\end{adjustbox}
\caption{Main results: per-language accuracy (\%) and macro-averages for closed, open-source, and cascaded systems, sorted chronologically by release date. Relevant modalities, $T$: Text and $A$: Audio are also specified.}
\label{tab:mainresults}
\end{table*}

\section{Results and Analysis}

\subsection{Main Results}

We present a comprehensive performance breakdown of all models on \textsc{GlobeAudio} in Table \ref{tab:mainresults}.

Evaluation on \textsc{GlobeAudio} reveals severe performance disparities across model families. \texttt{Gemini 3.1 Pro Preview} significantly outperforms the field at 82.68\% accuracy, whereas competing LALMs cluster at an average of 58.76\% and rarely surpass 70\%. This stark divide underscores a systemic vulnerability in current models: while they excel on structured benchmarks, they perform poorly when met with uncontrolled acoustic variability and compound reasoning demands inherent to naturalistic audio.

Overall, closed-source models consistently demonstrate stronger performance than open-source counterparts. Notably, \texttt{Gemini 2.0 Flash} outperforms most open-source LALMs despite being optimized for latency and cost. This persistent gap underscores the need for continued community efforts to advance open and reproducible LALMs.

Comparing all open-source LALMs, \texttt{Qwen3-Omni-Flash} substantially outperforms with an average accuracy of 72.11\%. This disparity may be influenced by its larger effective model capacity, as well as architectural choices such as its Mixture-of-Experts (MoE) design. Notably, the stronger closed-source models in our evaluation belong to the Gemini family, and also employ MoE-style architectures which suggests that expert routing may be beneficial for handling heterogeneous linguistic and acoustic conditions. This is further validated by \texttt{MiMo-V2.5}'s strong performance, which also employs such an architecture.

Performance further varies across languages. Higher-resource languages such as English and Russian generally yield stronger results across models, while low-resource languages such as Thai and Bengali remain more challenging, particularly for open-source LALMs and cascaded systems. Singlish, despite being a code-mixed variety, exhibits relatively strong performance across most models, with several systems achieving accuracies comparable to or higher than on English. This suggests that code-mixing alone does not necessarily degrade performance, especially when the dominant language is English.

\subsection{Audio Ablations and Analysis}

\begin{figure}
    \centering
    \includegraphics[width=\linewidth]{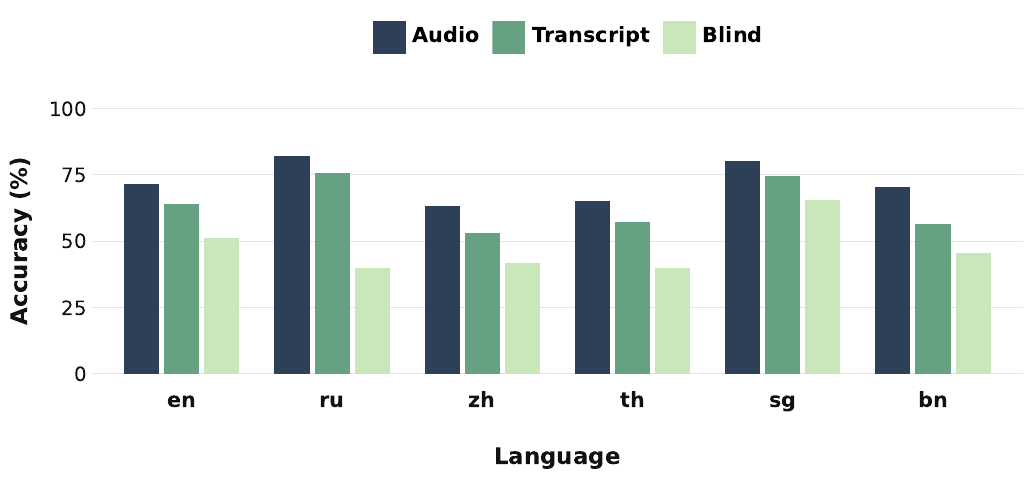}
    \caption{Accuracy of \texttt{Qwen3-Omni-Flash} on various settings, from Audio and Transcript input, to "Blind" which denotes model without audio or transcript input.}
    \label{fig:audio_transcript_blind}
\end{figure}

To systematically isolate the informational value of each modality and evaluate the models' causal reliance on context, we conduct ablations with the modality of the contextual input. Specifically, we evaluate the top-performing open-source model, \texttt{Qwen3-Omni-Flash}, across three distinct settings, while keeping the query in the source language:

\begin{itemize}[itemsep=2pt]
    \item \textit{Audio}: The model is given the raw acoustic signal to solve \textsc{GlobeAudio} questions, serving as our primary evaluation pipeline.
    \item \textit{Transcript}: To decouple linguistic semantics from acoustic nuances, the audio input is removed and reduced to its text transcript.
    \item \textit{Blind}: To detect guessing heuristics and reliance on prior world knowledge, the model is made to answer questions with zero supporting context, serving as a baseline.
\end{itemize}

\paragraph{Reliance on Audio}

When the model is subjected to the \textit{Blind} setting, we observe a sharp performance drop across all languages, with absolute accuracy falling to an average of 47.20\%, yielding a sharp average decrease of 24.91\%.

This sharp degradation demonstrates the close coupling between provided auditory contexts and \textsc{GlobeAudio}'s expertly crafted questions, and that crucially, internal world knowledge and query-side shortcuts are fundamentally insufficient to perform well without its audio samples.

\paragraph{Embedded prosodic and acoustic cues}

In the \textit{Transcript} setting, where the context is reduced strictly to text, we observe a marked and consistent performance decay across all linguistic families, resulting in an average absolute decrease of 8.69\% relative to the \textit{Audio} setting. Since the underlying semantic content remains identical between these two settings, this marked performance delta is likely attributed to the utility of the embedded prosodic, and paralinguistic cues native to our raw audio samples than to textual variance.

This validates that transcripts act as a lossy compression that flattens critical prosodic and paralinguistic cues. Furthermore, we directly demonstrate that \textsc{GlobeAudio} rigorously assesses an LALM's capacity to parse and understand messy, non-linguistic acoustic signals present in our dataset without reliance on linguistic patterns. However, we note that this degradation represents a conservative lower bound due to the presence of culturally grounded questions which inherently lean on text-retrievable context to answer.

\subsection{Cross-lingual Comprehension Analysis}

To investigate LALMs’ cross-lingual comprehension capabilities, we aim to assess the model's reliance on language-specific cues when processing complex auditory context.

\paragraph{Linguistic Fidelity}
To probe the depth of the linguistic cues present within \textsc{GlobeAudio}, we evaluate \texttt{Qwen3-Omni-Flash} using original source-language questions, comparing performance against English-translated counterparts. This allows us to quantify the information loss when language-specific nuances are subjected to translation.

We observe that their performance dipped slightly when subjected to English questions, by an average of 2.11\% with the exception of Bengali which performed better on English questions by 4.17\%. Despite framing the question in English, a high resource language, we discover that models still generally perform better in source languages.

This is possibly because \textsc{GlobeAudio}'s human-annotated questions contain dense, language-specific nuances that facilitate precise alignment with the provided natural audio. These nuances were likely irreversibly flattened during translation, depriving the model of critical grounding cues.
On the other hand, the models’ robust English capabilities might have mitigated the performance gap.

Conversely, the performance gain in Bengali, a relatively much lower resource language, suggests the model's possible reliance on English grounding when native-language depth is limited. This underscores the high-resource complexity of other language sets, where the original linguistic structure consistently outperforms translated alternatives.

\begin{figure}[h]
    \centering
    \includegraphics[width=\linewidth]{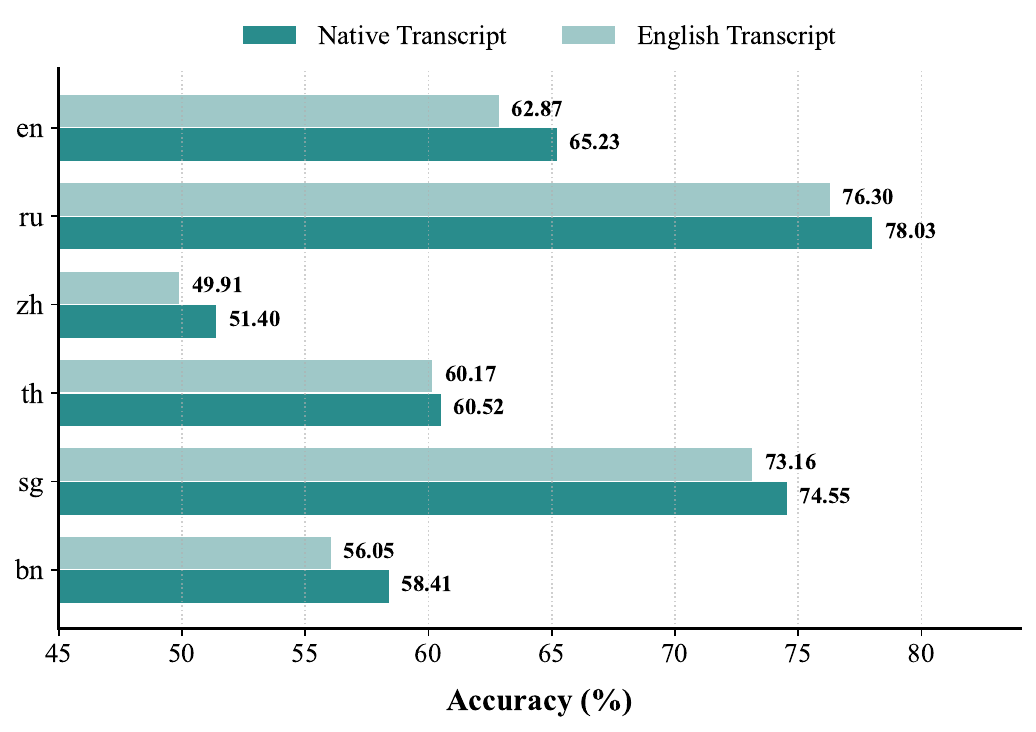}
    \caption{Analysis on varying language of transcripts in Cascade pipelines. We report average accuracy across all languages.}
    \label{fig:transcript-lang}
\end{figure}

\paragraph{Contextual Nuance}
To validate our findings by further analysing language-specific nuance present in \textsc{GlobeAudio}, we conduct a parallel experiment on the audio context and isolate linguistic from prosodic cues using Qwen3 LLM's cascade pipeline, which is particularly strong in multilingual understanding, and keep questions in their source language. We then compare its performance given source language transcripts with their English-translated counterparts and visualise results in Figure \ref{fig:transcript-lang}.

Concretely, we observe a similar consistent cross-lingual performance decline which validates our previous findings, and confirms that models struggle primarily with the loss of intrinsic linguistic and cultural structures in \textsc{GlobeAudio}. These elements are non-fungible across translations, suggesting that the performance gap is driven by a loss of semantic depth present in our dataset, rather than purely by the absence of prosodic information demonstrated by earlier ablations.

\section{Related Work}
\paragraph{Scattered Evaluation Efforts of LALMs}
The evaluations of LALMs has made considerable progress across diverse areas in recent years. IFEval-Audio \cite{gao2025ifevalaudiobenchmarkinginstructionfollowingcapability} is an audio benchmark assessing instruction-following across 6 diverse dimensions. AHa-Bench \cite{cheng2025ahabench} is a comprehensive benchmark for evaluating auditory hallucinations in LALMs to enhance their robustness and stability. SpeechR \cite{yang2025speechrbenchmarkspeechreasoning} is a speech-based benchmark assessing LALMs' reasoning capabilities. To assess complex reasoning across speech, environmental sounds and music, MMAU \cite{sakshi2024mmaumassivemultitaskaudio} and AIR-Bench \cite{yang2024airbenchbenchmarkinglargeaudiolanguage} was introduced. As a step towards evaluating under \textit{in-the-Wild} conditions, WildSpeech-Bench \cite{zhang2025wildspeechbenchbenchmarkingendtoendspeechllms} was introduced to assess auditory capabilities through synthetic mixtures of diverse speaker types and background noise. ProsodyEval \cite{yang2026measuringprosodydiversityzeroshot} is a benchmark introduced to assess ability of LALMs in capturing prosodic variations.

\paragraph{Universal Evaluation Efforts}
To unify such scattered evaluation efforts to keep up with the rising adoption of LALMs, universal benchmarks aim to provide wide coverage across a diverse range of tasks. AudioBench \cite{wang2025audiobenchuniversalbenchmarkaudio} was introduced to evaluate LALMs over 8 distinct tasks and 26 datasets, targeting 3 main aspects, namely speech understanding, audio scene understanding and voice understanding. On the other hand, MSEB \cite{heigold2025massive} was introduced to cover diverse categories of tasks, including Retrieval, Reranking, Reasoning and Classification.

\paragraph{Multilingual Evaluation}
Recognising the linguistic diversity, evaluation efforts have also begun to extend English-centric settings. Fleurs-SLU \cite{schmidt2025fleursslumassivelymultilingualbenchmark} examines 2 task types, namely speech classification across a wide range of languages. mSTEB \cite{beyene2025mstebmassivelymultilingualevaluation} assesses LALMs on 5 speech-based tasks and 5 text-based tasks, across a broad range of languages. Towards \textit{in-the-Wild} evaluation, ADU-Bench \cite{gao-etal-2025-benchmarking} is an audio dataset assessing 12 skills in 3 general scenarios, across 9 languages.

Despite these efforts, audio evaluation for LALMs remains substantially under-specified relative to real-world requirements, due to the lack of acoustic realism, and true linguistic and cultural authenticity.

\section{Conclusion}

In this work, we introduce \textsc{GlobeAudio}, a comprehensive high-fidelity audio benchmark designed to realistically evaluate LALMs under real-world conditions. Amidst the recent proliferation of audio evaluation suites, \textsc{GlobeAudio} establishes as a standardized, unified benchmark that embraces and rigorously examines LALMs' ability to handle the acoustic complexity and sociolinguistic diversity of in-the-wild contexts. Through our extensive experiments and analysis on state-of-the-art LALMs, we demonstrate that \textsc{GlobeAudio} contains multiple key characteristics including embedded prosodic and acoustic cues, and is well-equipped to rigorously examine the lingual and cultural understanding of LALMs. Ultimately, by highlighting the current vulnerabilities of these models, we establish a foundation for the community to build inclusive, highly performant LALMs suited for complex, real-world environments.

\section{Limitations}
\paragraph{Language Coverage}
In our work, we perform language selection to ensure coverage of diverse language families, geographic regions, and ensuring substantial variation in resource availability. However, we acknowledge this might not fully capture all linguistic diversity or account for the full range of resource conditions across languages. As such, we plan to expand \textsc{GlobeAudio}'s language coverage to better capture global linguistic diversity.

\paragraph{Data distribution}
The resulting distribution after data sourcing is inherently shaped by platform-side filtering. Since unavailable or private videos cannot be indexed, the distribution of our obtained data might be skewed towards content that adhere to platform guidelines, potentially underrepresenting more controversial sources and media.

To mitigate the latter's impact, we (1) use the \texttt{tv\_simply} player client for better reach and (2) query across a wide set of keyword templates.


\appendix
\section{Appendix}
\label{sec:appendix}

\subsection{Prompts used in Inference}
\label{sec: appendix-prompts}
\paragraph{Overview}
Across all experiments, we employ a unified prompting template to ensure consistency and comparability across models. Specifically, each model is prompted with the instruction:
\textit{"Please answer this multiple-choice question: <question>"}. followed by the four answer options labeled \textit{A}, \textit{B}, \textit{C}, and \textit{D}. To constrain model outputs and avoid verbose generations, we append the explicit instruction:
\textit{“Do NOT include any explanation. ONLY output a single character: A, B, C, or D.”}
Finally, we include the cue \textit{“Answer:”} to elicit the model’s response for evaluation.

\begin{figure*}[h]
    \centering
    \includegraphics[width=\linewidth]{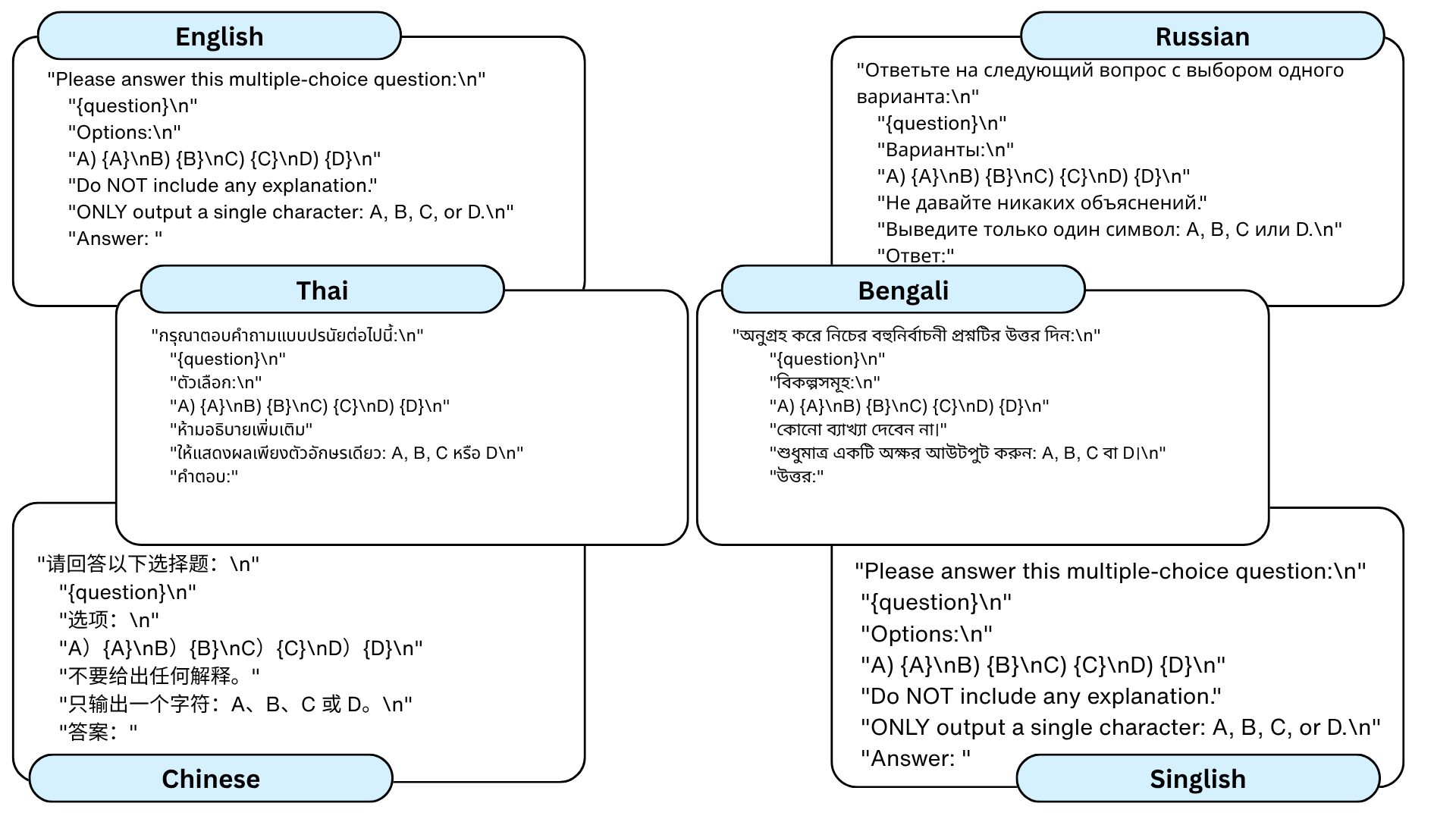}
    \caption{Prompts for Open and Closed-Source Models in various languages}
    \label{fig:prompts-appendix}
\end{figure*}

\paragraph{Prompts for Open and Closed-Source Models}
Refer to Figure \ref{fig:prompts-appendix}.

\begin{figure*}[h]
    \centering
    \includegraphics[width=\linewidth]{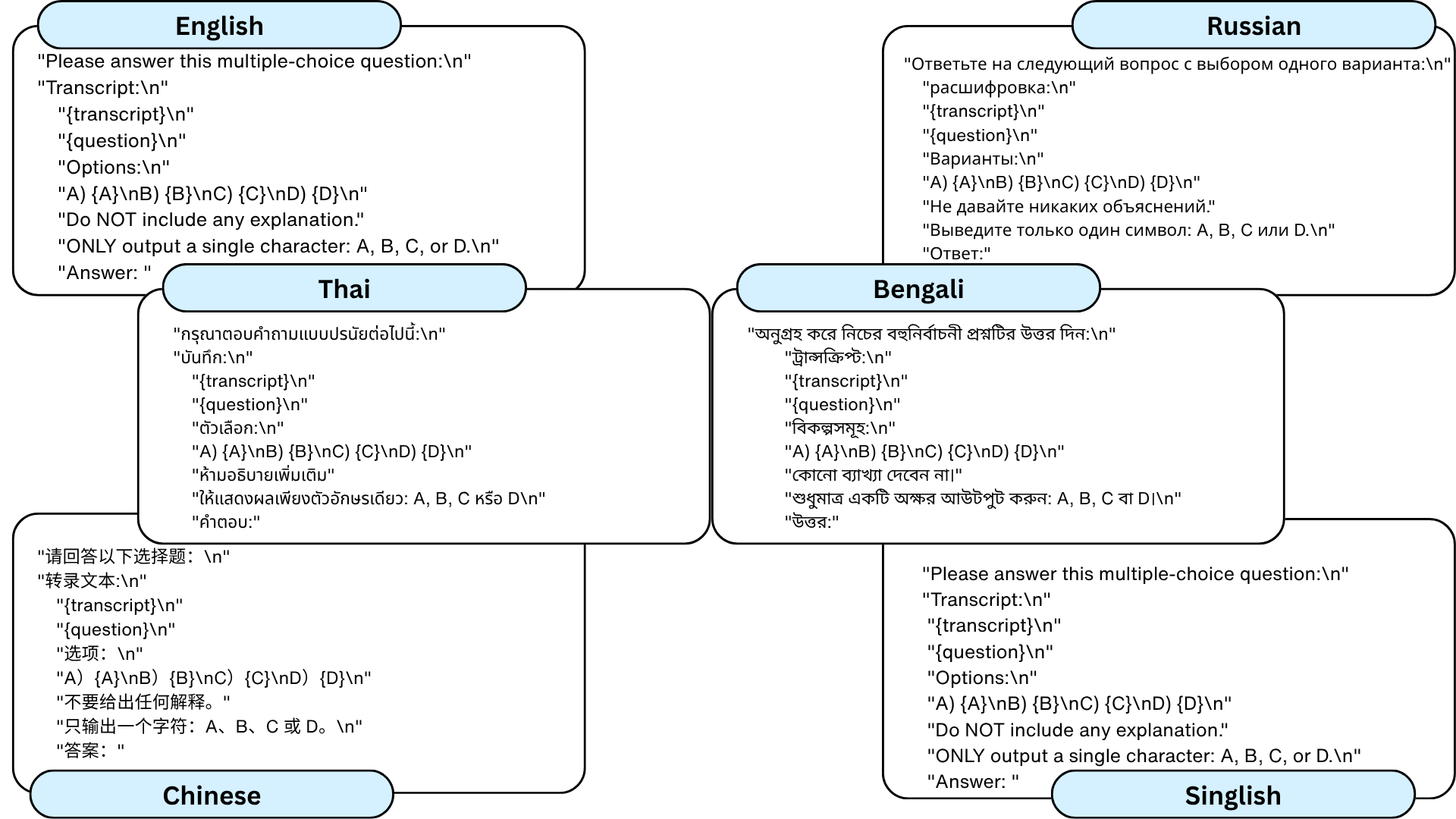}
    \caption{Prompts for Cascade pipelines in various languages}
    \label{fig:cascade-prompts-appendix}
\end{figure*}

\paragraph{Prompts for Cascade Pipelines}
Refer to Figure \ref{fig:cascade-prompts-appendix}.

\subsection{\textsc{GlobeAudio}'s Question Examples of each language}
\label{sec:appendix-eg}
\begin{figure}[H]
    \includegraphics[width=\columnwidth]{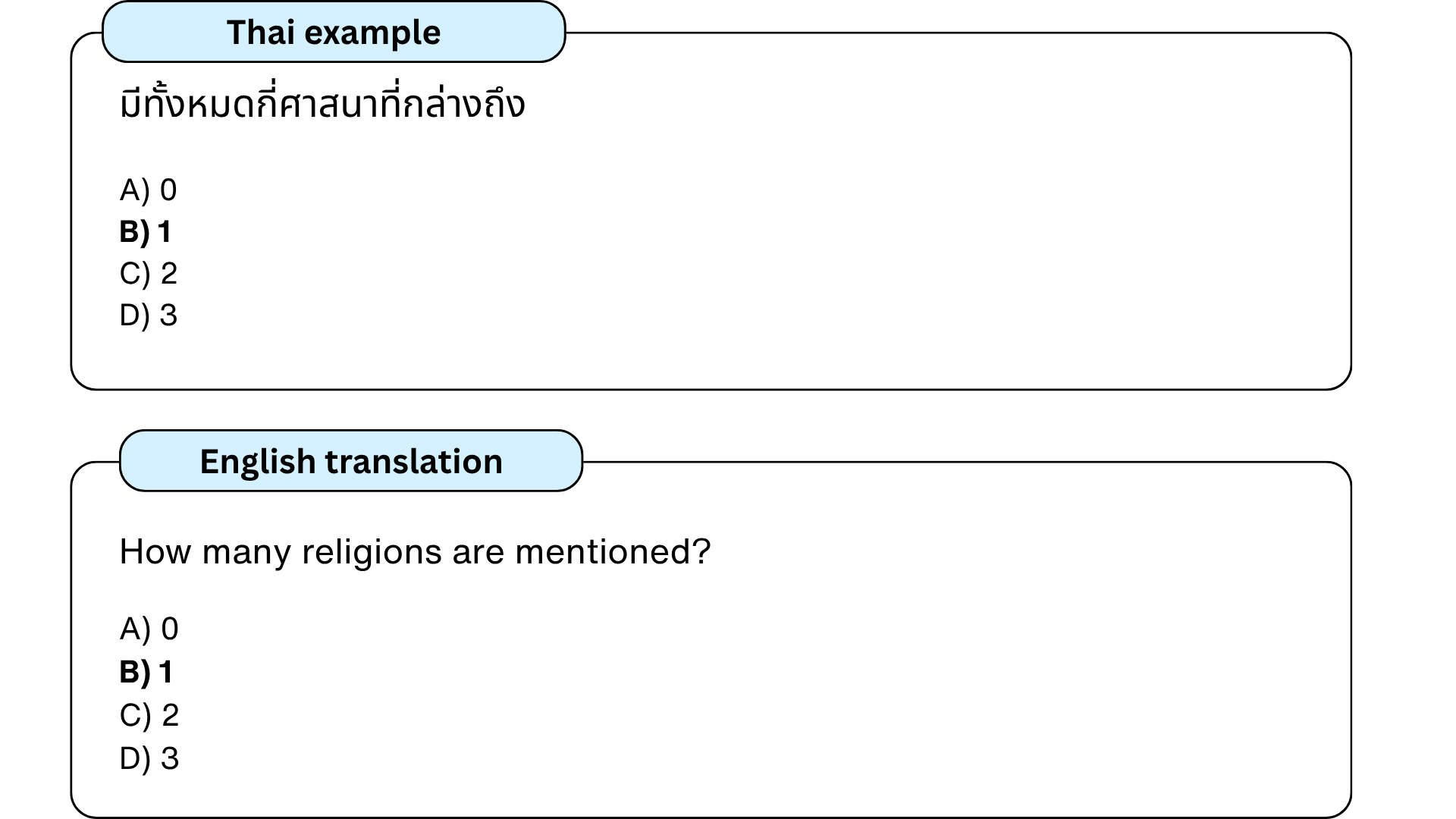}
    \caption{An example of a Thai question from \textsc{GlobeAudio}}
    \label{fig:thai-example}
\end{figure}

\begin{figure}[H]
    \includegraphics[width=\columnwidth]{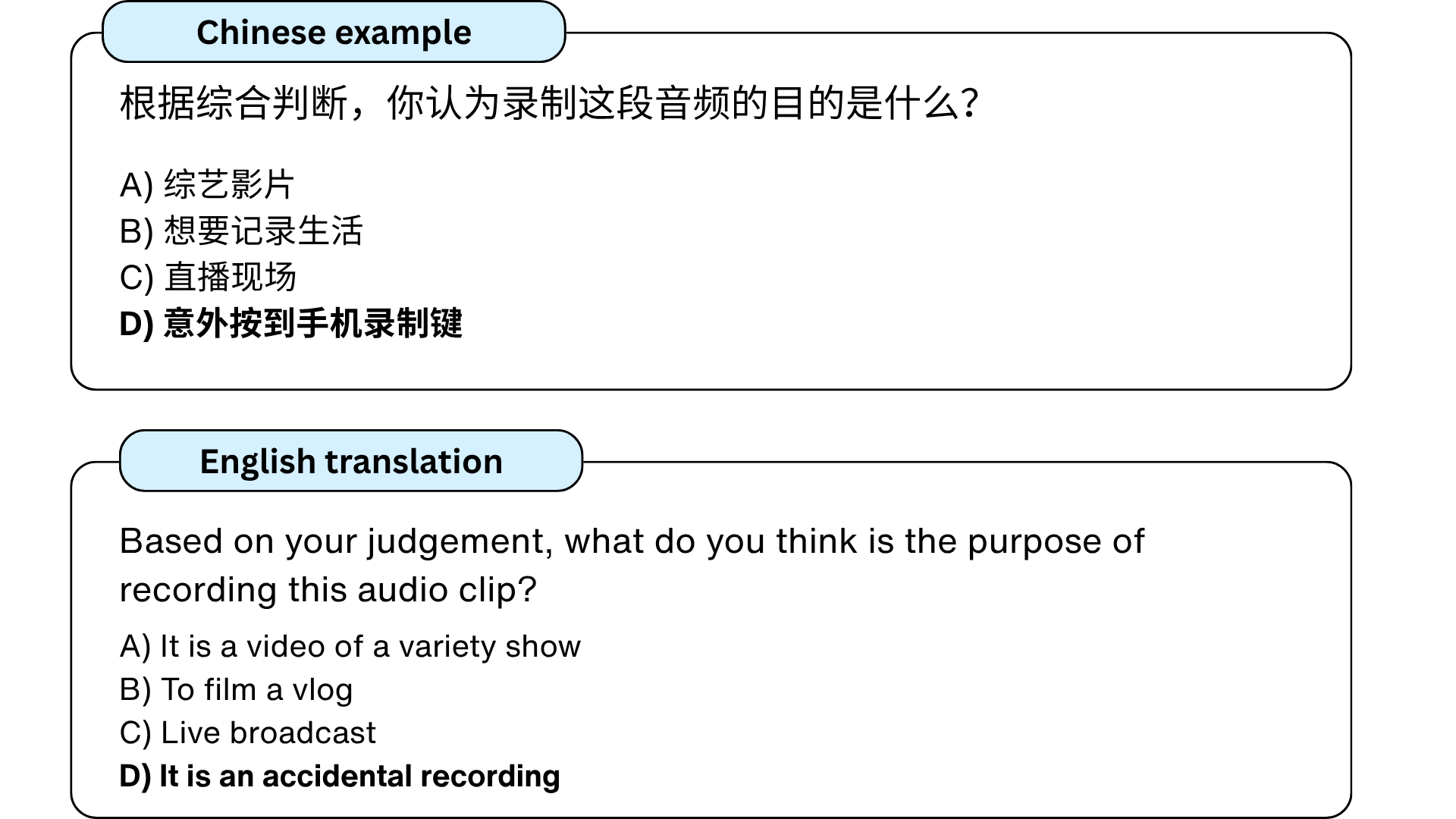}
    \caption{An example of a Chinese question from \textsc{GlobeAudio}}
    \label{fig:chinese-example}

\end{figure}

\begin{figure}[H]
    \includegraphics[width=\columnwidth]{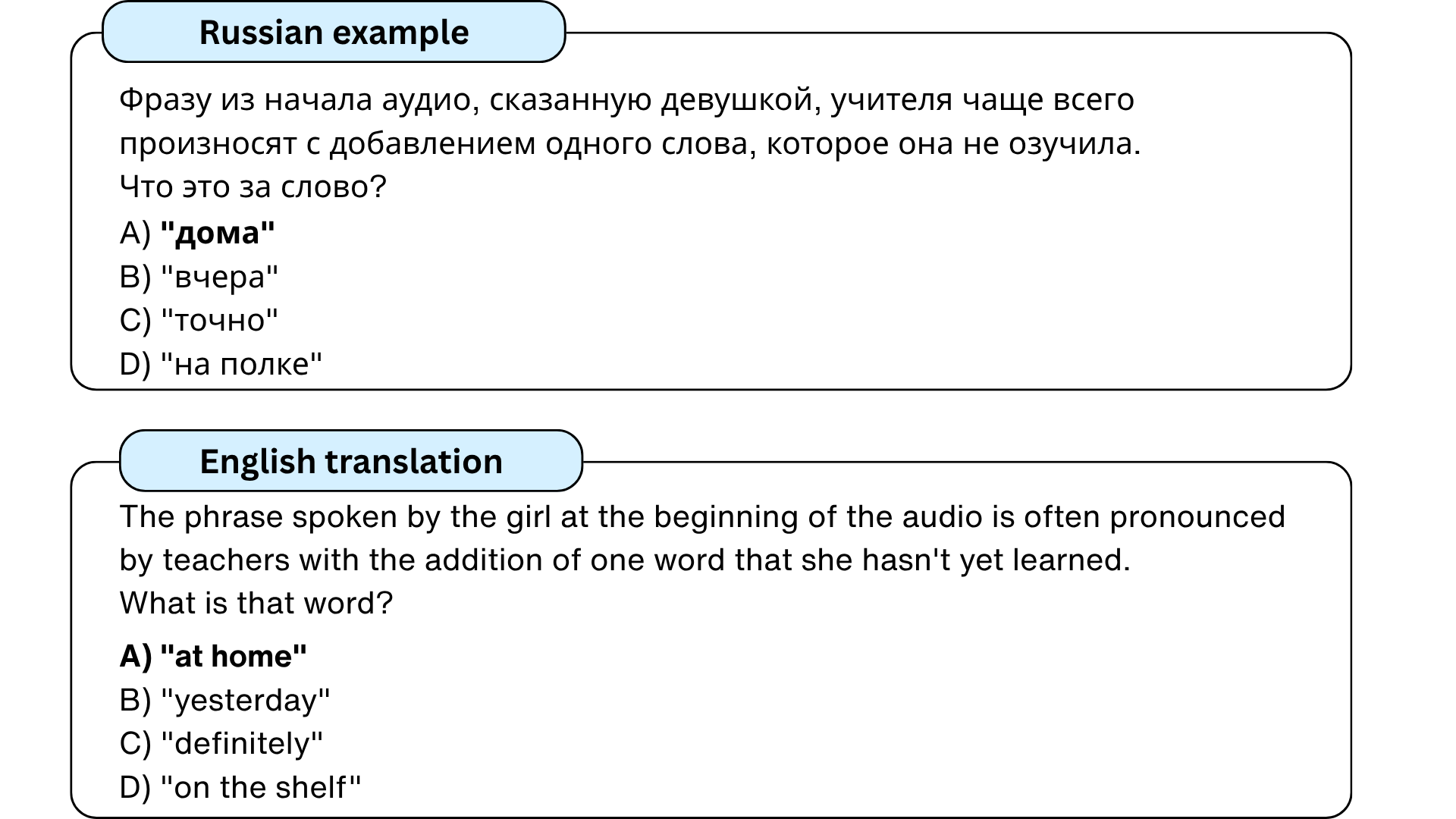}
    \caption{An example of a Russian question from \textsc{GlobeAudio}}
    \label{fig:russian-example}

\end{figure}

\begin{figure}[H]
    \includegraphics[width=\columnwidth]{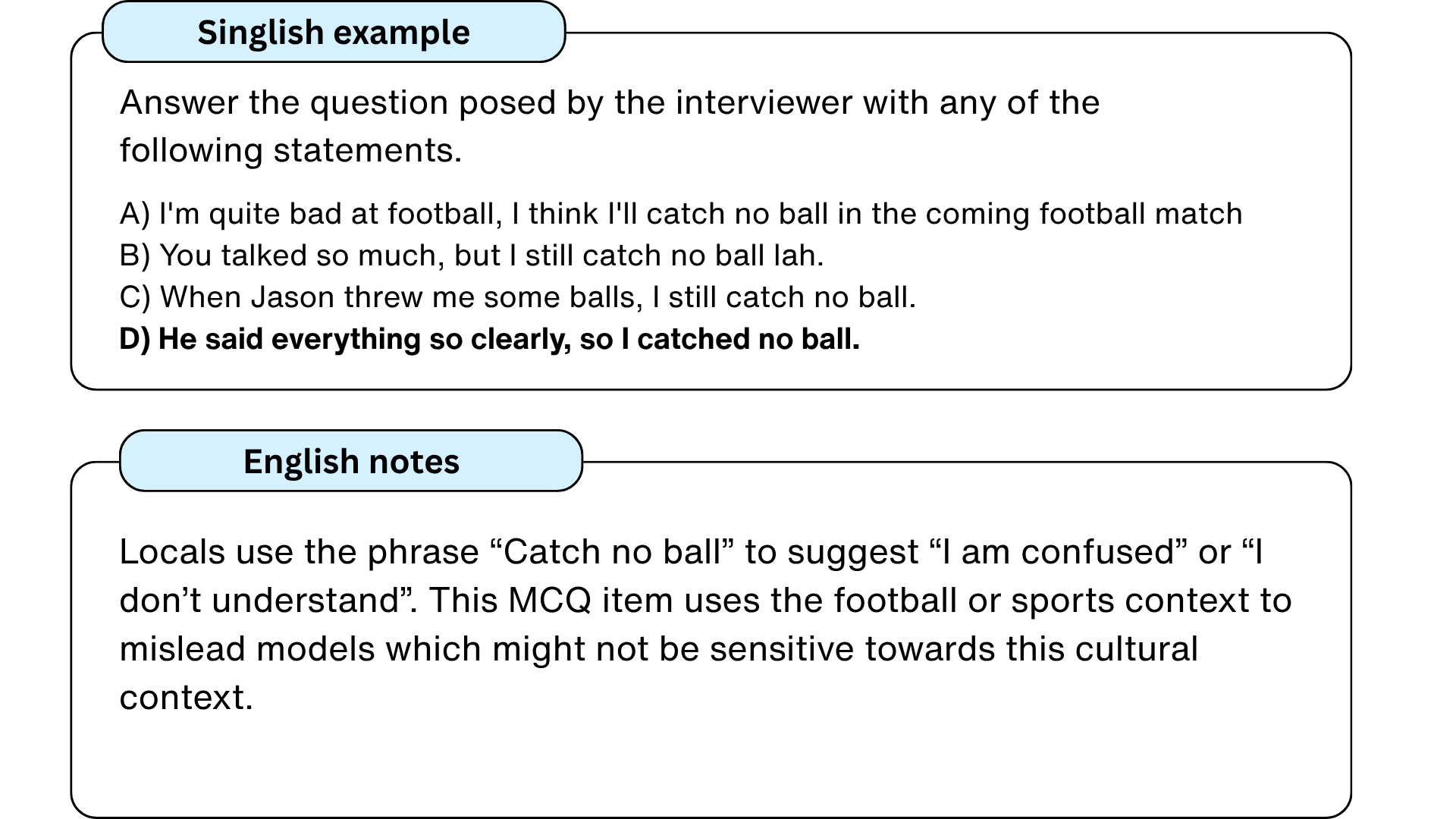}
    \caption{An example of a Singlish question from \textsc{GlobeAudio}}
    \label{fig:singlish-example}

\end{figure}

\begin{figure}[H]
    \includegraphics[width=\columnwidth]{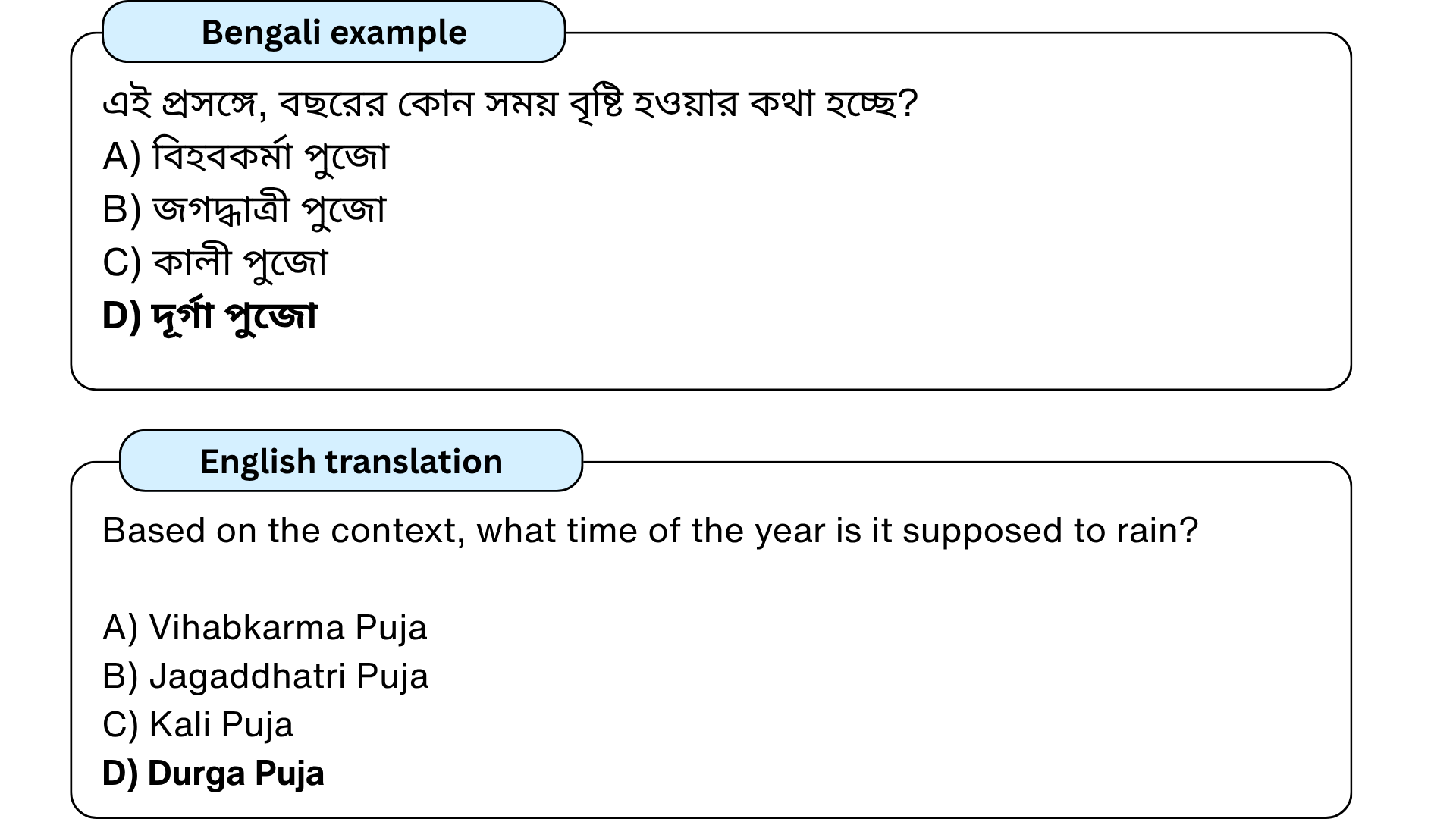}
    \caption{An example of a Bengali question from \textsc{GlobeAudio}}
    \label{fig:bengali_example}

\end{figure}

\subsection{\texttt{yt-dlp} Commands and Parameters}
\begin{figure*}
    \centering
    \includegraphics[width=\linewidth]{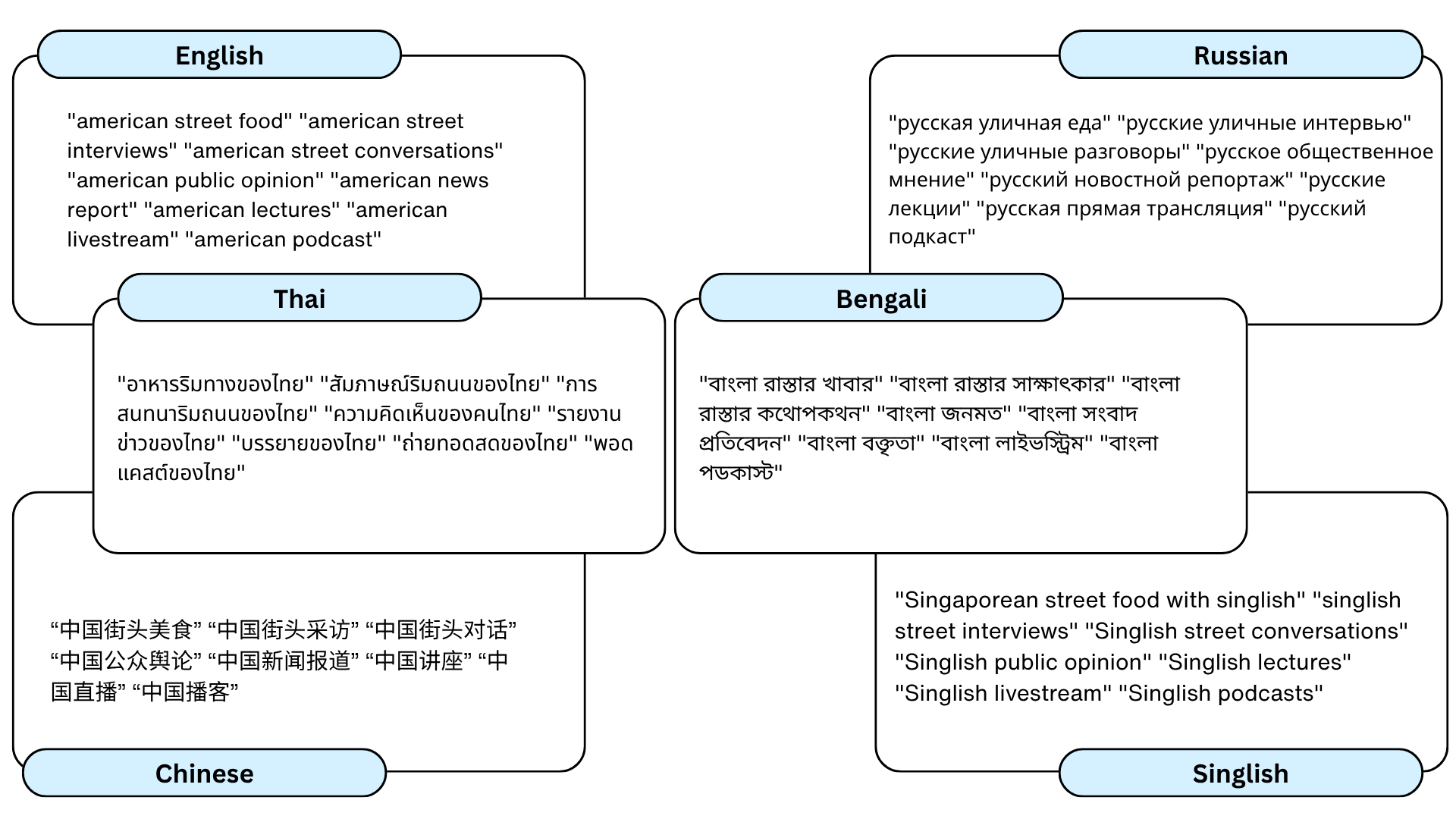}
    \caption{YouTube downloader query terms}
    \label{fig:search-queries}
\end{figure*}

Our query terms are detailed in Figure \ref{fig:search-queries}. 

We use the following commands on \texttt{yt-dlp} to source our audio clips from YouTube. For each query term as detailed in Figure \ref{fig:search-queries}, we run:

yt-dlp "ytsearch15:\${term}" --get-id | sed 's\verb|_^_|https://www.youtube.com/watch?v=\_' >> <lang>\_urls.txt

Then, we run this command to extract the audio from the URLs gathered in the .txt file and ensure we download the version with the best audio quality:

<lang>\_urls.txt | xargs -n 1 -P 8 yt-dlp -x --extractor-args "youtube:player-client=default,-tv\_simply" --audio-format mp3 --embed-metadata --add-metadata --write-info-json -f bestaudio -o "\%(title)s.\%(ext)s"

\end{document}